\documentclass[letterpaper, 10 pt, conference]{ieeeconf}  
\IEEEoverridecommandlockouts                              
\usepackage[T1]{fontenc}
\usepackage{amsmath,amsfonts}
\usepackage{algorithmic}
\usepackage[ruled,linesnumbered]{algorithm2e}
\usepackage{array}
\usepackage{bm}
\usepackage{textcomp}
\usepackage{stfloats}
\usepackage{url}
\usepackage{verbatim}
\usepackage{graphicx}
\usepackage{tabularx,booktabs}
\usepackage{multirow}
\usepackage{multicol}
\usepackage{colortbl} 
\usepackage{xcolor}
\newcommand{\bb}{\boldsymbol}

\usepackage{amssymb}
\usepackage{bbding}
\usepackage{pifont}
\usepackage{colortbl}
\DeclareMathOperator{\diag}{diag}

\definecolor{myblue}{RGB}{42, 73, 161}
\definecolor{myred}{RGB}{212, 45, 42}

\usepackage{hyperref} 
\hypersetup{
    colorlinks = true,   
    pdfborder = {0 0 0}, 
    linkcolor  = myred,
    citecolor  = myblue,
    urlcolor   = magenta
}

\makeatletter
\let\ieee@citex\@citex
\def\@citex[#1]#2{%
  \hyper@linkstart{cite}{cite.#2}%
  \ieee@citex[#1]{#2}%
  \hyper@linkend
}
\makeatother
\let\labelindent\undefined
\usepackage{enumitem}

\title{\LARGE \bf
Assistance Torque Estimation via Dynamics-Aware Optimization for Lower-Limb Exoskeleton in Complex Environments
}

\author{Xiao-Yin Liu$^{1,2}$, Guotao Li$^{1,2\text{ }*}$, Weiqun Wang$^{1,2}$ and Zeng-Guang Hou$^{1,2\text{ }*}$
\thanks{*Corresponding Authors (guotao.li@ia.ac.cn, zengguang.hou@ia.ac.cn).}
\thanks{$^{1}$State Key Laboratory of Multimodal Artificial Intelligence Systems, Institute of Automation, Chinese Academy of Sciences, Beijing, China.}%
\thanks{$^{2}$The School of Artificial Intelligence, University of Chinese Academy of Sciences, Beijing, China.}%
}

\begin{document}

\maketitle
\thispagestyle{empty}
\pagestyle{empty}

\begin{abstract}
Ground-truth human joint torque estimation relies on motion capture systems, which suffer from limited outdoor usability and significant deployment expenses. Furthermore, direct scaling of ground-truth joint torques to obtain motor torque commands is not necessarily the optimal strategy. To address the aforementioned limitations, inspired by the human motion generation process, this paper proposes a novel assistance torque estimation method based on the dynamic model. 
From an optimization perspective, the proposed method directly generates motor‑assist torque and lowers the cost of data acquisition.
Then, a data-driven assistance torque prediction network is trained to enable accurate real-time prediction under complex outdoor environments. Experimental results demonstrate that optimized (estimated) assistance torque exhibits better phase consistency with gait trajectories and better alignment with task characteristics.
Relative to the Zero torque condition, the predicted torque can decrease metabolic rate by $11.8\% \sim 17.7\%$, heart rate by $8.9\% \sim 14.3\%$, and peak muscle activation levels by $28.2\%\sim54.0\%$, respectively. This provides a new perspective for low-cost adaptive exoskeleton assistance. Related Website: \href{https://youtu.be/CGxDD0jKpak}{Movie}.
\end{abstract}

    \section{Introduction}
The core of human-robot interaction lies in the robot's torque output matching human motion intentions \cite{slade2022personalizing,liu2025weight}. However, human joint torque cannot be measured directly and is most often inferred from high-cost motion capture systems (expensive plantar force plates) \cite{molinaro2024task}, which limits the range of practical application scenarios for robots. To enable the practical deployment of exoskeletons in complex unstructured environments, a low-cost, readily deployable data acquisition method is critically important \cite{luo2024experiment}. Extensive research has been conducted on torque estimation methods in the existing literature.

Methods for joint torque estimation can be broadly classified into two categories: inverse dynamics (ID)-based approaches \cite{molinaro2024task,2023id,2022id,diraneyya2021inertial,molinaro2024estimating} and electromyography (EMG)-based approaches \cite{quesada2026emg,2019emg,emg2021}. The first type of method, ID-based approaches, relies on measurements of human kinematic parameters and ground reaction forces \cite{2023id}. Kinematic data can be captured via motion capture systems or inertial measurement units (IMUs), while ground reaction forces are typically measured using foot pressure sensors \cite{2022id}. While serving as the gold standard for ground-truth torque measurement \cite{molinaro2024task}, these methods incur high deployment costs and have limited utility in practical assistance scenarios.
For the second type of methods, EMG-based approaches \cite{2019emg} establish different mathematical models between EMG and torque to estimate torque. However, these methods are highly susceptible to inter-subject variability and external disturbances \cite{quesada2026emg,bird2026ultra}, which severely limit their applicability for torque estimation in highly dynamic scenarios \cite{emg2021}.

\begin{figure}
	\centering
	\includegraphics[width=0.46\textwidth]{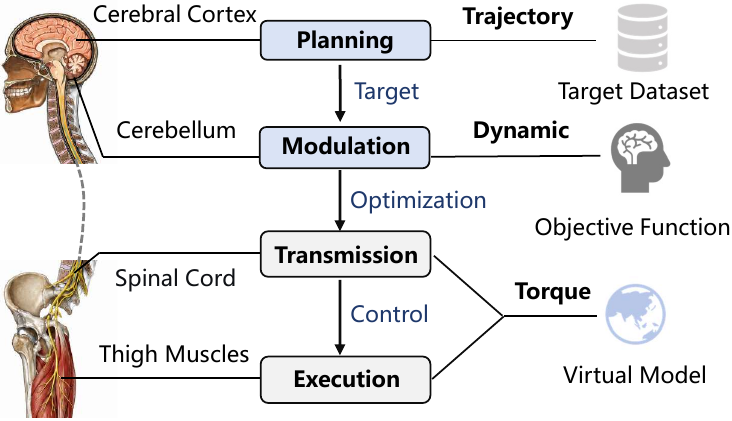}
	\caption{The schematic depicts motion generation: biologically, the cerebral cortex produces desired trajectories, while the cerebellum optimizes motor commands via an internal dynamics model and relays them to muscles via the spinal cord; for the exoskeleton, the optimization objective is built from reference trajectories and its dynamics model, and then assistance torque is tuned by the objective and virtual environment feedback.}. 
    \vspace{-0.4cm}
	\label{fig1}
\end{figure}

To address the limitations of existing torque estimation methods for exoskeleton assistance, this paper adopts a novel optimization-based perspective on torque estimation. This research perspective draws inspiration from the physiological process of human motion generation \cite{grillner2003motor, miall1993cerebellum}: 1) The cerebral cortex generates motor intentions, high-level locomotor plans, gait targets, and motor trajectories in a top-down control hierarchy \cite{scott2004optimal};
2) The cerebellum receives descending cortical motor commands. It uses stored internal dynamic models to compare desired motion with real-time sensory feedback and fine-tune the timing and magnitude of muscle activation \cite{miall1993cerebellum}. The detailed workflow is illustrated in Fig. \ref{fig1}. Essentially, the cerebral cortex conducts high-level motor planning, while the cerebellum generates control commands via internal dynamic models and modulates muscle activation to execute planned motor tasks. 
The underlying principle of the aforementioned motor control process bears conceptual consistency with that of model predictive control \cite{wolpert2000computational}.

The exoskeleton system is inherently a human–robot coupled system. Although the dynamic models of the human lower limb and the hip exoskeleton share a consistent formulation and are coupled via the interaction torque, their dynamic parameters differ \cite{YAN2022439,liu2026personalized}, and precise alignment between the exoskeleton joints and the human joints during motion remains difficult to achieve \cite{10716507,9798696}. Consequently, under the condition that the exoskeleton conforms to human motion, the motor output torque and the human joint torque do not exhibit a simple linear relationship \cite{liu2026exotraj}. Since the collected joint data correspond to the exoskeleton motor side, this paper performs direct optimization of the motor assistance torque using the exoskeleton dynamics model based on the process of human motion generation. Specifically, motor angles and angular velocities measured during human walking with the exoskeleton are designated as desired trajectories, while kinematic states iteratively computed via the dynamic model serve as actual trajectories, which are generated by the virtual model. The motor assistance torques are then optimized to minimize the tracking error between the actual and desired trajectories.

\begin{figure*}
	\centering
	\includegraphics[width=1.0\textwidth]{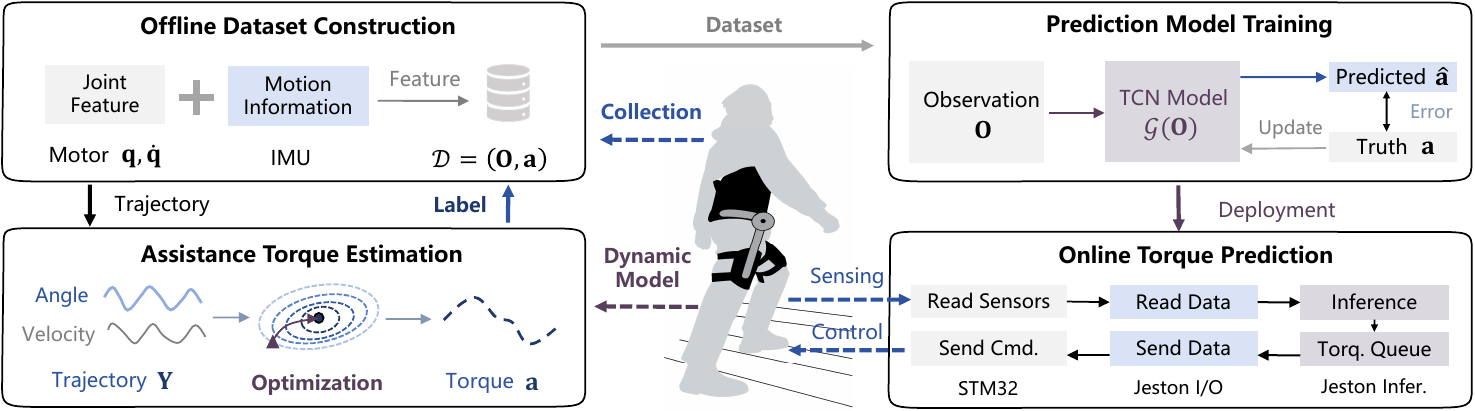}
	\caption{Overall framework of the proposed torque prediction system. Motor trajectories $\mathbf{Y}$ are collected, and corresponding torques $\mathbf{a}$ estimated via dynamics optimization are set as dataset labels. Acquired motor and IMU measurements are taken as input features $\mathbf{O}$ to build an offline dataset $\mathcal{D}$. A torque prediction model $\mathcal{G}$ is trained on the dataset and deployed on hardware, where the microcontroller performs real-time sensor reading and transmits torque values inferred by the Jetson Orin NX edge computing unit.}. 
    \vspace{-0.4cm}
	\label{fig2}
\end{figure*}

To achieve adaptive exoskeleton assistance in outdoor complex environments, this paper synchronously acquires multi-modal sensory data, including motion states measured by IMUs, motor angles, and angular velocities, from human subjects walking with the exoskeleton in complex unstructured scenarios. Then, leveraging the aforementioned torque optimization method, this paper derives an optimized motor-assistance torque, which is subsequently used as ground-truth labels for training the torque prediction model. In the offline phase, the proposed assistance torque prediction model adopts a temporal convolutional network (TCN) backbone. It is trained by iteratively updating the parameters to minimize the prediction error against ground-truth labels, generating the final prediction model. Fig. \ref{fig2} presents a schematic illustration of the overall framework of the proposed method.

The efficacy of the proposed method is validated in unstructured outdoor environments. Experimental results demonstrate that the optimized motor assistance torque maintains better gait phase consistency with gait trajectories and can adapt to different task characteristics. The torque prediction model yields a coefficient of determination (\(R^2\)) of $0.84$. Relative to the zero torque baseline condition, the peak of muscle activation, metabolic rate, and heart rate are reduced by $28.2\%\sim54.0\%$, $11.8\% \sim 17.7\%$, and $8.9\%\sim14.3\%$, respectively. These findings corroborate that the proposed torque estimation approach effectively reduces human metabolic energy expenditure and enhances assistance performance in dynamic outdoor scenarios, without incurring prohibitive data acquisition overhead. The key contributions of this work are outlined below:

\begin{itemize}
    \item A novel exoskeleton assistance torque estimation method is proposed, which is based on dynamic model optimization and requires no costly data acquisition.
    \item The proposed prediction model built upon the estimated torque can adapt to dynamic environments and effectively reduce human energy expenditure.
\end{itemize}

This work opens up a new perspective for exoskeleton assistance in complex environments, which can significantly reduce the reliance on expensive hardware platforms.

    \section{Related Works}
This section presents methodologies for exoskeleton torque estimation and strategies for exoskeleton assistance in complex dynamic environments.

\subsection{Torque Estimation}
In human–robot interaction, existing torque estimation methods fall into two main categories: inverse dynamics-based \cite{molinaro2024estimating,molinaro2024task, 2023id,2022id} and EMG-based  \cite{quesada2026emg,2019emg,emg2021} approaches. The first category of methods estimates joint torques from limb kinematic data and plantar pressure measurements using musculoskeletal models \cite{molinaro2024estimating,molinaro2024task} or mathematical models \cite{2023id,2022id}. However, this category of methods requires motion capture systems or IMUs, as well as costly plantar pressure plates, for data acquisition, incurring prohibitive instrumentation costs and constraining viable data collection scenarios. 
The second category of methods estimates joint torques from measured EMG signals via Hill-type models \cite{2019emg}, nonlinear parametric regression models \cite{quesada2026emg}, or neural network models \cite{emg2021}. However, this category of methods is hampered by inter-individual variability (necessitating subject-specific parameter calibration) and limited EMG signal fidelity, resulting in torque estimation accuracy that remains inadequate for real-world exoskeleton assistance applications. This paper proposes an optimization-driven exoskeleton assistance torque estimation approach that requires no costly data acquisition instruments and features strong robustness to ambient noise, rendering it well-suited for adaptive assistance in complex outdoor environments.

\subsection{Assistance in High-Dynamic Environment}
To date, two primary methodological frameworks have been established for exoskeleton assistance in highly dynamic environments: torque estimation-based approaches \cite{molinaro2024estimating,molinaro2024task} and methods grounded in integrated musculoskeletal–exoskeleton models \cite{luo2024experiment}.
Methods in the first category primarily estimate joint torques via motion capture systems (MoCaps), and they inherit the aforementioned limitations: prohibitive data acquisition costs and restricted applicability in complex outdoor settings.
The second category trains reinforcement learning control policies within simulation environments, eliminating the need for experimental data collection. However, due to the high dimensionality of musculoskeletal (MSK) systems, developing a control policy adaptable to diverse complex terrains remains challenging, which in turn limits the generalizability across unstructured terrain scenarios.
The method proposed in this work falls into the torque estimation paradigm (estimating assistance torque instead of joint torque) and enables efficient exoskeleton assistance in complex terrains while satisfying low-cost deployment constraints.

\begin{table}
	\normalsize
	\caption{Comparison results of state-of-the-art methods for exoskeleton assistance in high-dynamic scenarios.}
	\label{tab_1}
	\centering
    \vspace{-0.1cm}
	\resizebox{8.6cm}{!}{
		\begin{tabular}{c|cc|cc}
\toprule[1pt] 
\textbf{Related Works} & \textbf{MoCaps} & \textbf{MSK} & \textbf{Data Cost} & \textbf{Adaptive}\\

\midrule
Molinaro et al. (2024) \cite{molinaro2024estimating}& \Checkmark & \XSolidBrush & High &Low\\
Molinaro et al. (2024) \cite{molinaro2024task}&\Checkmark& \XSolidBrush  & High &High\\
Luo et al. (2024) \cite{luo2024experiment}&\XSolidBrush& \Checkmark & Low &Low \\
\midrule
Ours &\XSolidBrush& \XSolidBrush & Low &High \\
\bottomrule[1pt]
\end{tabular}
}
\vspace{-0.3cm}
\end{table}
	\section{Torque Estimation via Dynamic-Aware Optimization}
This section delineates the methodological pipeline of dynamics-model-based torque optimization (estimation) without expensive data collection costs for complex environments.

\subsection{Dynamic Model}
Assuming the human lower limb is modeled as a rigid-body system, the lower limb and exoskeleton subsystems are coupled through mechanical interactions at the wearable attachment interfaces. The resulting human-exoskeleton coupled dynamic model can be formulated as follows \cite{YAN2022439}:
\begin{equation}\label{eq_1}
\begin{aligned}
\mathbf{M}_e\ddot{\mathbf{q}}_e+\mathbf{C}_e(\mathbf{q}_e,\dot{\mathbf{q}}_e)\dot{\mathbf{q}}_e+\mathbf{G}_e(\mathbf{q}_e)&=\mathbf{T}_{e}-\mathbf{T}_{int},\\
\mathbf{M}_h\ddot{\mathbf{q}}_h+\mathbf{C}_h(\mathbf{q}_h,\dot{\mathbf{q}}_h)\dot{\mathbf{q}}_h+\mathbf{G}_h(\mathbf{q}_h)&=\mathbf{T}_{h}+\mathbf{T}_{int},
\end{aligned}
\end{equation}
where subscripts $e$ and $h$ denote the exoskeleton and the human body, respectively. \(\mathbf{M}\), \(\mathbf{C}\), and \(\mathbf{G}\) represent the mass matrix, Coriolis-centrifugal terms, and gravity term, respectively. $\mathbf{q}_e$ and $\mathbf{q}_h$ are the angles of the motor and human hip joint, and $\mathbf{T}_{e}$ and $\mathbf{T}_{h}$ denote the motor output torque and the human hip joint torque, respectively. $\mathbf{T}_{int}$ denotes the torque of human-robot interaction, which can be estimated via the impedance model:
\begin{equation}
    \mathbf{T}_{int}=\mathbf{K}_p(\mathbf{q}_h-\mathbf{q}_e)+\mathbf{K}_d(\dot{\mathbf{q}}_h-\dot{\mathbf{q}}_e),
\end{equation}
where $\mathbf{K}_p$ and $\mathbf{K}_d$ denote the stiffness matrix and damping matrix, respectively. However, in complex unstructured locomotion scenarios, human joint angles are not directly measurable, and accurate estimation of human-robot interaction torques remains a key challenge. When predicted trajectories are used as surrogates for human joint trajectories, the inherent perception-control coupling tends to degrade the stability of exoskeleton assistance.

When exoskeleton assistance is compliant with human joint torques, human-robot interaction torques act in the opposite direction to human joint torques, and the absolute magnitude of the interaction torques increases with joint angular velocity. Accordingly, the human-robot interaction torques estimated from joint angular velocities can be formulated as follows:
\begin{equation}\label{eq_3}
    \mathbf{T}_{int} = \mathbf{K}_1 \dot{\mathbf{q}}_e+\mathbf{T}_0,
\end{equation}
where $\mathbf{K}_1$ and $\mathbf{T}_0$ are coefficients. Accordingly, with the prescribed reference trajectories \((\mathbf{q}_e^d, \dot{\mathbf{q}}_e^d)\) as inputs, the optimal motor assistance torque $\mathbf{T}_{e}$ can be derived via the optimization formulation established in Eqs. \eqref{eq_1} and \eqref{eq_3}. In particular, dynamic parameter mismatches between human lower limbs and the exoskeleton preclude a trivial linear mapping between motor assistance and human joint torques. Furthermore, in this work, the motor assistance torque optimization procedure is simplified by exclusively accounting for the dynamic model of the exoskeleton.

\subsection{Torque Optimization based on Dynamic Model}
For notational simplicity, the left subscript $e$ denoting exoskeleton dynamic model parameters is omitted throughout the subsequent derivations.
Define the state vector as \(\mathbf{y}=[\mathbf{q};\dot{\mathbf{q}}]\in \mathbb{R}^{4}\) and the control input torque as \(\mathbf{T}\in \mathbb{R}^{2}\). Based on Eqs. \eqref{eq_1} and \eqref{eq_3}, the governing dynamic relation linking the state time derivative \(\dot{\mathbf{y}}\) to the control torque \(\mathbf{T}\) is expressed as
\begin{equation}
    \mathbf{\dot{y}}=\begin{bmatrix} \mathbf{y}_1 \\ \mathbf{M}^{-1}\big(\mathbf{T-\mathbf{C}}\mathbf{y}_1+\mathbf{G}(\mathbf{y}_0)-\mathbf{T}_{int}(\mathbf{y}_1)\big) \\ \end{bmatrix},
\end{equation}
where $\mathbf{y}_0=\mathbf{q}$ and $\mathbf{y}_1=\dot{\mathbf{q}}$. In the discrete-time system, the next state can be denoted as $\mathbf{y}_{t+1}=\mathbf{y}_{t}+\mathbf{\dot{y}}_t\Delta t$. Then, the state \(\mathbf{y}_{t}\) and control torque \(\mathbf{T}_t\) satisfy a deterministic transition function f mapping the current state-action pair to the next state:
\begin{equation}
    \mathbf{y}_{t+1}=f(\mathbf{y}_{t},\mathbf{T}_t).
\end{equation}

During natural human locomotion, gait trajectories are initially planned by the cerebral cortex \cite{wolpert2000computational}. In this work, motor angles $\mathbf{q}$ and angular velocities $\dot{\mathbf{q}}$ measured from human subjects during natural walking with the exoskeleton are designated as the desired trajectories. Subsequently, based on the planned trajectories, the cerebellum computes motor commands via its internal dynamic models \cite{scott2004optimal}. Essentially, this control logic can be formulated as an optimization problem. 
In this work, our method is built upon the exoskeleton dynamic model and aims to optimize motor assistance torques. Given the desired trajectory \(\mathbf{Y}^d = [\mathbf{y}_1;...;\mathbf{y}_L]\in\mathbf{R}^{4L}\) and the optimized assistance torque \(\mathbf{A} = [\mathbf{T}_1;...;\mathbf{T}_L]\in\mathbf{R}^{2L}\), where $L$ denotes the length of the trajectory, the optimization objective function can be expressed as follows:
\begin{equation}\label{eq_6}
    \begin{cases}
        \arg\min\limits_{\mathbf{A}}\text{ } &\left(\mathbf{Y}^d-\mathbf{{Y}}\right) ^{\mathbf{T}}\mathbf{Q}\left(\mathbf{Y}^d-\mathbf{{Y}}\right)
        + \mathbf{A}^{\mathbf{T}}\mathbf{P} \mathbf{A},\\
        \text{ s.t. }  & \mathbf{y}_{t+1}=f(\mathbf{y}_{t},\mathbf{T}_t), 
    \end{cases}
\end{equation}
where $\mathbf{Q}\in \mathbf{R}^{4L\times4L}=\diag\{\mathbf{Q}_1,..., \mathbf{Q}_{L}\}$ is the error weight matrix and $\mathbf{P}\in \mathbf{R}^{2L\times2L}=\diag\{\mathbf{P}_1,..., \mathbf{P}_{L}\}$ is the torque constrain weight matrix. $\mathbf{Q}_{1:L-1}\in \mathbf{R}^{4\times4}=\diag\{C_q\mathbf{I}_2,C_v\mathbf{I}_2\}$, $\mathbf{Q}_L=C_p\mathbf{I}_4$ and $\mathbf{P}_{1:L}=C_a\mathbf{I}_4$. $C_q$, $C_v$, $C_p$, and $C_a$ denote the weighting coefficients for angular error, angular velocity error, terminal error, and torque penalty term, respectively. Term $\mathbf{A}^{\mathbf{T}}\mathbf{P} \mathbf{A}$ serves as a regularization term in the objective function, preventing the torque from becoming excessively large. 
Using the collected continuous trajectory data, the optimization is performed via a sliding-window scheme with window length $L$ and sliding step size $1$. The first torque value of $\mathbf{A}$ solved from each window-wise optimization is regarded as the estimated assistance torque corresponding to the current motion state.
	\section{Adaptive Assistance via Torque Prediction}
The estimation method for the motor assistance torque has been presented above. To achieve adaptive exoskeleton assistance, a torque prediction model should be trained, that is, a model that predicts the future torque $\mathbf{a}$ from past observations $\mathbf{O}$.

\subsection{Offline Dataset Construction}
This paper collects human motion data in complex environments using only $18$-dimensional IMU measurements (from three IMUs) and $4$-dimensional motor measurements (from two motors). This sensor configuration obviates the need for expensive motion capture systems, thereby substantially reducing the cost of data acquisition. Subsequently, based on the assistance torque estimation method presented above, the torque values are calculated from the acquired $4$-dimensional motor data, \textit{i.e.}, the angles and angular velocities of the left and right motors, and are then used as the ground-truth labels for the torque prediction network.

Subsequently, the offline training dataset can be constructed for the assistance torque prediction model. The input features consist of the 18‑dimensional IMU measurements and the angular positions of the left and right motors, while the corresponding $2$‑dimensional torque values serve as the ground‑truth labels. Specifically, at each time instant $t$, the observation $\mathbf{O}_t=[{\mathbf{o}_{t-T_o-1},...,\mathbf{o}_{t}}]\in \mathbf{R}^{20\times T_o}$ is defined as the sequence of input features spanning the past $T_o$ time steps, and the target $\mathbf{a}_t=\mathbf{T}_{t+T_p}\in \mathbf{R}^{2}$ is the torque at $T_p$ time steps into the future. From this formulation, the offline dataset 
$\mathcal{D}=\{\mathbf{O}_t,\mathbf{a}_t\}_{i=1}^N$ can be constructed, where $N$ is the number of sample.

\subsection{Torque Prediction Model Training}
Based on the offline dataset $\mathcal{D}$, this section focuses on learning a mapping function $\mathcal{G}$ from observations $\mathbf{O}_t$ to torques $\mathbf{a}_t$, that is,
\begin{equation}
    \mathbf{a}_t=\mathcal{G}(\mathbf{O}_t).
\end{equation}
Due to the complexity of the observations, a neural network parametrized with $\omega$ is employed to approximate the mapping $\mathcal{G}$. Accordingly, the optimization objective for updating the network parameters can be formulated as
\begin{equation}
    \arg\min_{\omega} \text{ } \mathcal{L}=\mathbb{E}_{\mathbf{O},\mathbf{a}\sim \mathcal{D}}\big[\|\mathcal{G}_{\omega}(\mathbf{O})-\mathbf{a}\|_2\big].
\end{equation}
To capture the temporal dependencies and inter-feature correlations inherent in the observations $\mathbf{O}$, a temporal convolutional network (TCN) is adopted to parameterize the mapping function $\mathcal{G}(\cdot)$, with the network weights collectively denoted by $\omega$. Once trained, the torque prediction model $\mathcal{G}$ is converted to the TensorRT format and deployed on an edge computing device, thus enabling real-time inference and real-time torque assistance for the exoskeleton.
	\section{Experiments}
\subsection{Experimental setup}
\begin{figure}
	\centering
	\includegraphics[width=0.48\textwidth]{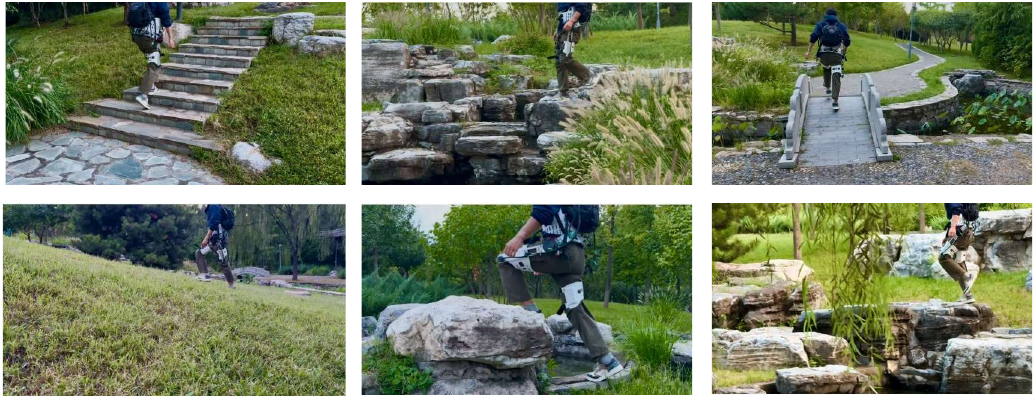}
	\caption{Schematic of outdoor assistance locomotion terrain over rocky, sloped, and grassy terrains.}. 
    \vspace{-0.6cm}
	\label{fig3}
\end{figure}

The experiments are divided into an offline phase and an online phase. In the offline phase, the assistance torque is derived from the collected motor angles and angular velocities, and the TCN model is trained to predict the future motor torque output. In the online phase, the assistance torque is predicted in real time. The user’s muscle activation level, metabolic cost, and heart rate are then compared across three conditions: Zero torque (wearing the exoskeleton without assistance), Assist on (wearing the exoskeleton with assistance), and No exo (not wearing the exoskeleton).

Before the experiment, all participants were fully informed of the experimental procedures. They were then instructed to walk on a treadmill, on stairs, and over complex outdoor terrain under the three conditions. Fig. \ref{fig3} presents a schematic diagram of the complex outdoor scenario, which comprises various outdoor terrains including lawns, steep slopes, and rocky formations. The experimental section of this work is designed to address two fundamental questions: 
\begin{itemize}[leftmargin=*] 
    \item $\mathbf{Q}_1:$ Does the assistance torque estimated from the dynamics model conform to the human motion intention? 
    \item $\mathbf{Q}_2:$ What are the torque prediction performance and assistance effectiveness in complex environments?
\end{itemize}


\begin{figure*}
	\centering
	\includegraphics[width=1.0\textwidth]{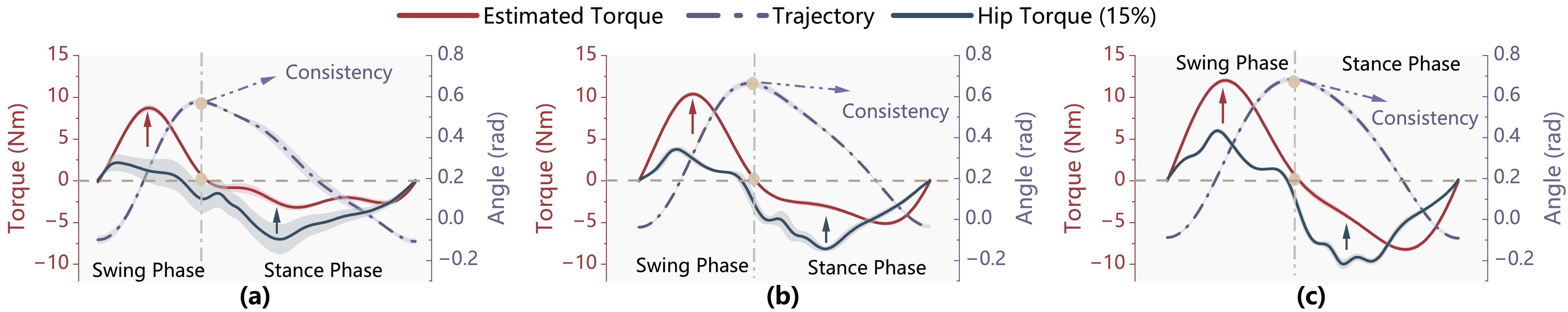}
	\caption{The relationship results between the estimated torque and joint trajectories, as well as their correspondence with measurements from the motion capture system. Panels (a), (b), and (c) present the experimental results at walking speeds of 0.6, 1.0, and 1.4 m/s, respectively.}
    \vspace{-0.0cm}
	\label{fig4}
\end{figure*}

\begin{figure*}
	\centering
	\includegraphics[width=1.0\textwidth]{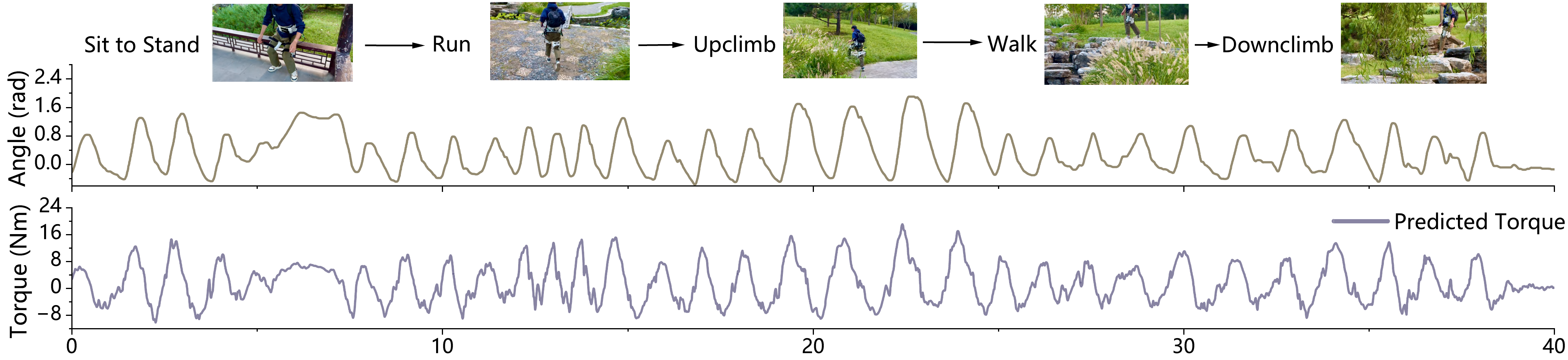}
	\caption{The results of the real-time torque prediction of the proposed method and related trajectory under the outdoor complex task conditions.}
    \vspace{-0.4cm}
	\label{fig5}
\end{figure*}

\subsection{Assistance Torque Estimation Results (\texorpdfstring{$Q_1$}{Q1})}
\subsubsection{Torque Estimation Results}
The quality of the estimated torque is assessed through two complementary dimensions: 1) kinematic consistency with the gait trajectory; 2) compatibility with task characteristics.
The first dimension evaluates the temporal coherence between the estimated torque profile and gait kinematics. Specifically, it verifies whether the estimated torque converges to zero as the gait trajectory approaches its peak values during the stance and swing phases.
The second dimension validates the alignment between the estimated torque and inherent task properties. It primarily examines whether peak torque values change in response to dynamic task scenarios, such as changes in locomotion speed.

Fig. \ref{fig4} presents the collected motor trajectories and corresponding estimated torque under three representative locomotion conditions, namely level walking at speeds of $0.6$ m/s, $1.0$ m/s, and $1.4$ m/s. Across all three locomotion tasks, the estimated torque consistently converges to near-zero magnitudes at the phase transition points between stance and swing. This observation validates the kinematic consistency between the estimated torque profiles and the corresponding gait trajectories.
Furthermore, the peak magnitude of the estimated torque increases correspondingly with rising locomotion speed. These results collectively confirm that the estimated torque aligns well with the inherent characteristics of different locomotion tasks. 

\subsubsection{Analysis of Assistance Torque} This part presents a comparative analysis between the estimated hip joint torques and the reference hip joint torques computed via OpenSim using measurements from a motion capture system. Fig. \ref{fig4} presents the comparison between the torque estimated by the proposed method and the reference hip joint torque computed from a motion capture system, under level walking conditions at speeds of $0.6$ m/s, $1.0$ m/s, and $1.4$ m/s. The reference torque data are sourced from a public dataset \cite{camargo2021comprehensive}. Specifically, we select data from subjects with body height and mass comparable to those of the test participant in this work, and scale the reference torque values by a factor of $0.15$. This scaling is justified by the approximate proportional relationship of $0.15$ between the assistive torque and the total torque of the hip joint \cite{molinaro2024task}.


Experimental results reveal similar overall trends but significant amplitude differences between the scaled human joint torque and the optimized assistance torque. Human joint torque exhibits a pronounced stance-phase peak due to body-weight support, whereas the thigh-mounted exoskeleton cannot directly provide postural or weight-bearing assistance, making such high torque unnecessary. These findings indicate that proportional scaling of human joint torque is not an optimal strategy for generating assistance torque. Instead, the relationship between human joint torque and optimal assistance torque is nonlinear, and its underlying coupling mechanism deserves further investigation.

\subsection{Assistance Performance Results (\texorpdfstring{$Q_2$}{Q2})}
\subsubsection{Torque Prediction Results} 
All experiments presented below adopt a cross-subject evaluation protocol, where the feature data of test subjects remain completely unseen by the network throughout the entire training phase.
Fig. \ref{fig5} shows the predicted torque results and corresponding trajectories for the subject performing a range of locomotion tasks: walking, sit-to-stand, variable-speed running, and rock climbing. It shows that, under the complex terrain conditions, the predicted torque agrees well with the gait trajectories: torque zero-crossings correspond to stance-swing phase transitions, and torque magnitude rises with increasing speed, consistent with the physical properties of locomotion tasks.

Fig. \ref{fig9} summarizes the online prediction accuracy (the squared R ($R^2$)) and online inference latency (ms) of different models. The online torque label is calculated by Eq. \eqref{eq_6} based on collected angle and angle velocity. The experimental results indicate that TCN attains the best performance. Therefore, TCN is adopted in this work for online torque prediction. LSTM exhibits the poorest prediction performance, as its inference latency on edge devices (about $8$ ms) fails to meet the $5$ ms real-time constraint, and it incurs substantial prediction errors. TCN and Transformer achieve only moderate accuracy, limited by the cross-subject generalization requirement and a sensor feature distribution shift: the offline dataset is collected under no-load motor conditions, whereas loaded motor operation during real-time assistance alters the feature statistics. However, TCN maintains an overall consistent prediction trend, with deviations only at peaks and troughs, which still satisfies the requirements for exoskeleton assistance.



\begin{figure*}
	\centering
	\includegraphics[width=1.0\textwidth]{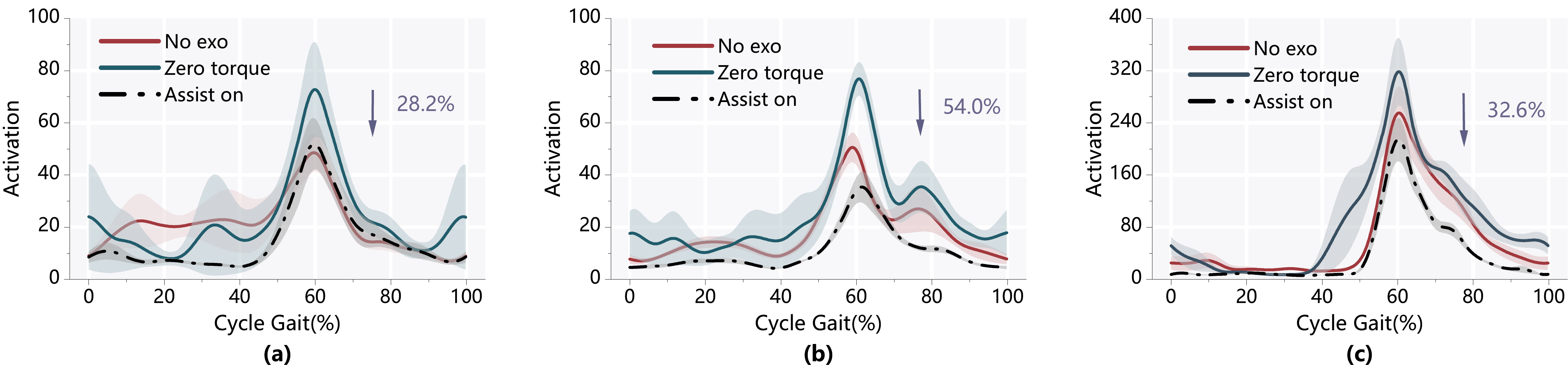}
	\caption{Comparison of muscle activation curves within one gait cycle across three assistance conditions for level-ground, ramp, and stair walking tasks.}
        \vspace{-0.4cm}
	\label{fig6}
\end{figure*}



\begin{figure}
	\centering
	\includegraphics[width=0.465\textwidth]{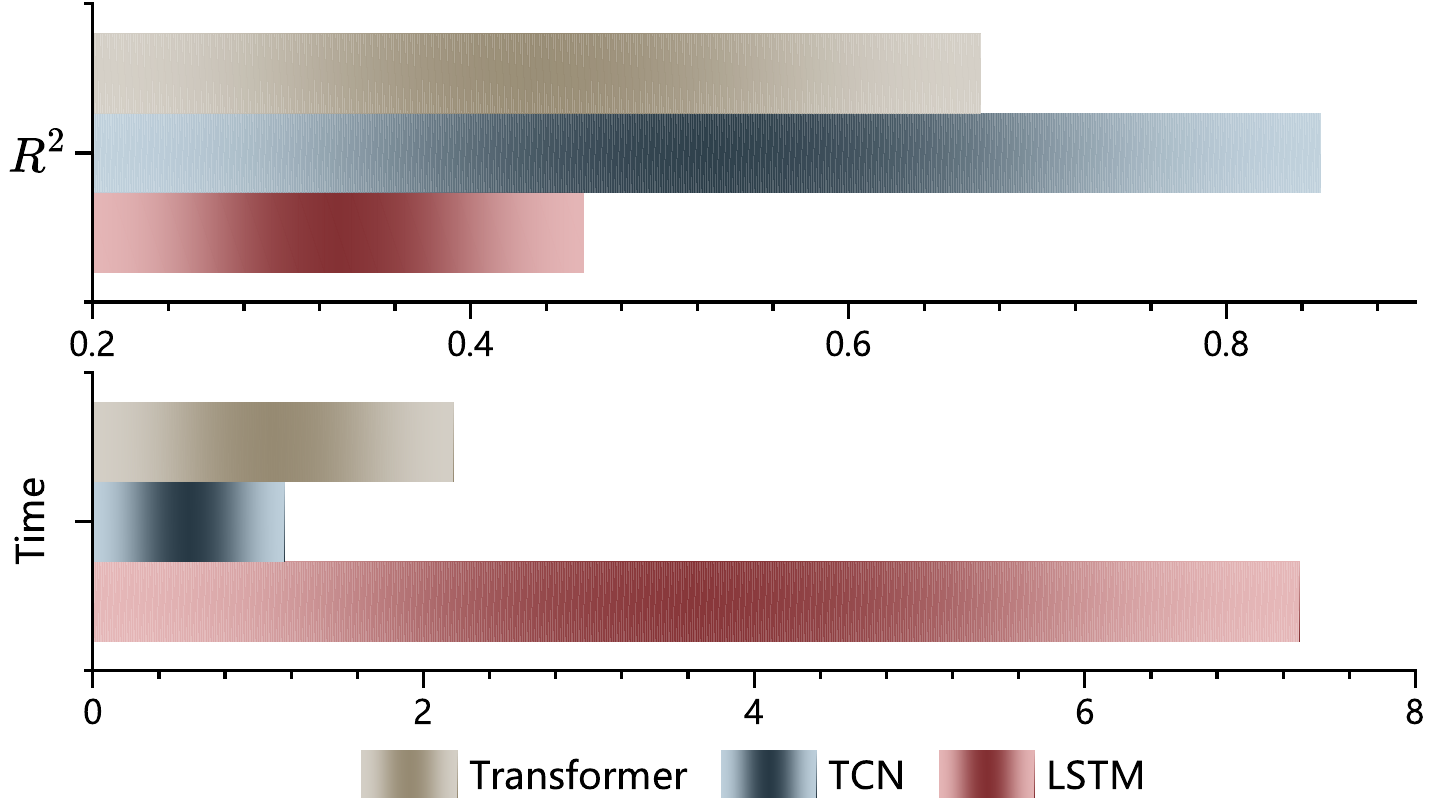}
	\caption{Comparison of online torque prediction errors and inference times across different network models.}. 
    \vspace{-0.6cm}
	\label{fig9}
\end{figure}

\subsubsection{Muscle Activation Results}
Given that EMG signals are highly sensitive to locomotion patterns, muscle activation was measured under three distinct assistance conditions across three locomotor tasks: level treadmill walking at $1.0$ m/s, $5$° inclined treadmill walking at $1.0$ m/s, and stair climbing, to characterize the modulation of muscular activation induced by different assistance strategies.
Fig. \ref{fig6} illustrates the muscle activation over a single gait cycle for each of the three tasks under all three assistance conditions, \textit{i.e.}, Assist on, Zero torque, and No exo. These results demonstrate that 1) Compare to Zero torque condition, assistance torque cuts peak muscle activation by $28.2\%$, $54.0\%$, and $32.6\%$ under the three terrain conditions;
2) Against the No exo condition, the peak muscle activation is reduced $-7.0\%$, $39.8\%$, and $16.1\%$ for the three respective terrains;
3) Relative to the Zero torque condition, integrated muscle activation is reduced by $30.2\%$, $41.0\%$, and $37.6\%$ for the three terrains in sequence.
The results show that the estimated torque from the optimization-based approach can effectively reduce muscle activation.

\begin{figure}
	\centering
	\includegraphics[width=0.5\textwidth]{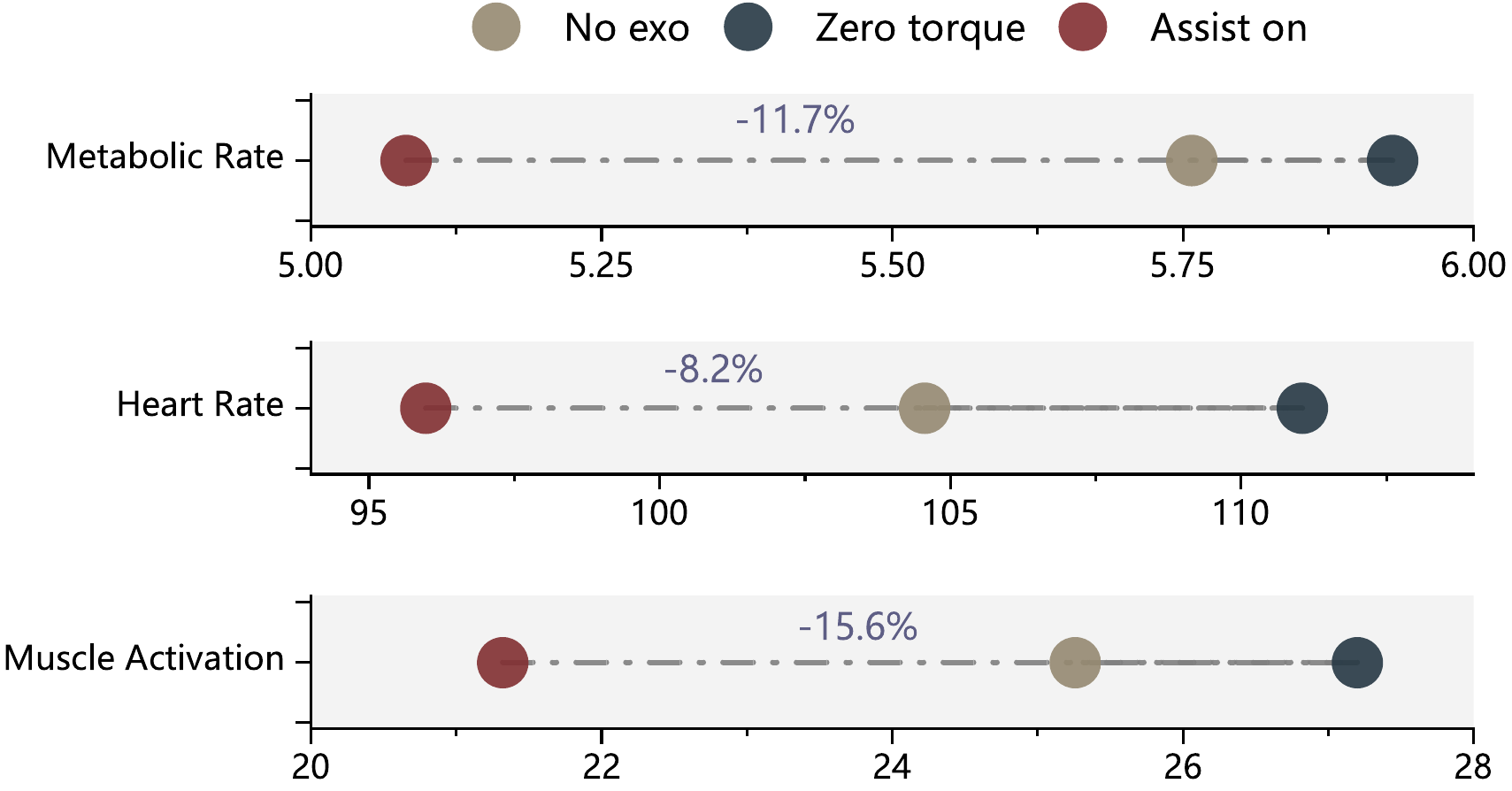}
	\caption{Comparison of metabolic cost, heart rate, and muscle activation across three assistance conditions on outdoor complex terrain.}
    \vspace{-0.6cm}
	\label{fig8}
\end{figure}

\begin{figure*}
	\centering
	\includegraphics[width=0.98\textwidth]{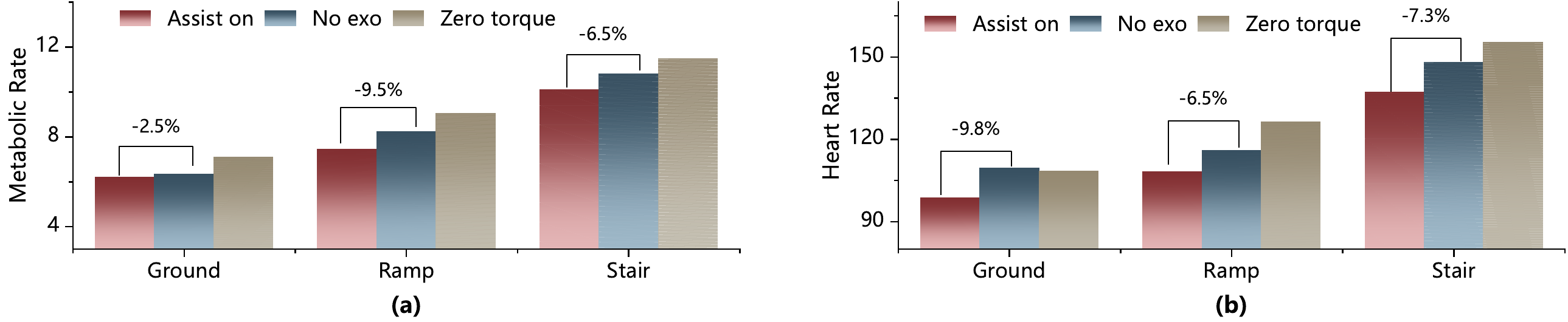}
    \vspace{-0.4cm}
	\caption{The comparison results of metabolic cost (a) and heart rate (b) under three assistance conditions during ground, ramp, and stair walking tasks.}
    \vspace{-0.6cm}
	\label{fig7}
\end{figure*}

\subsubsection{Metabolic and Heart Rate Results}
The experiments presented in this section are conducted under two distinct scenarios. The first is a complex, dynamic unstructured outdoor scenario, covering diverse locomotion tasks including rock scrambling, running, and walking on uneven grassy terrain. The entire outdoor test course spans approximately $1$ km and takes roughly $8$ minutes to traverse. The second is a well-controlled structured scenario comprising three canonical locomotion tasks: level walking, inclined walking, and stair climbing.
Fig. \ref{fig8} presents a quantitative comparison of metabolic cost, heart rate, and average muscle activation under the three assistance conditions in the unstructured outdoor scenario. The results show that, compared with the No exo condition, the predicted torque reduces metabolic cost, heart rate, and average muscle activation by $11.7\%$, $8.2\%$, and $15.6\%$, respectively. And, compared with the Zero torque condition, the three metrics are reduced by $14.3\%$, $13.6\%$, and $21.6\%$, respectively.

Fig. \ref{fig7} presents a quantitative comparison of metabolic cost and heart rate under all three assistance conditions across the three structured locomotion tasks. The results indicate that, relative to the Zero torque condition, metabolic cost decreases by $12.7\%$, $17.7\%$, and $11.8\%$ across the three tasks, with corresponding heart rate reductions of $8.9\%$, $14.3\%$, and $11.7\%$, respectively. Compared with the No exo baseline, metabolic cost decreases by $2.5\%$, $9.5\%$, and $6.5\%$ across the task set, with corresponding heart rate reductions of $9.8\%$, $6.5\%$, and $7.3\%$, respectively.  Collectively, experimental results show that the estimated torque effectively reduces users’ energy expenditure across a range of locomotion tasks.

    \section{Conclusion}
This paper proposes a novel optimization-based method to estimate exoskeleton assistance torque, effectively reducing data acquisition costs.
This estimation method optimizes the torque by incorporating the exoskeleton dynamics model.
Based on the estimated torque, an assistive torque prediction model is trained to achieve real-time assistance in complex environments. Compared with the Zero torque condition, the predicted assistance torque reduces metabolic rate by $11.8\% \sim 17.7\%$, heart rate by $8.9\% \sim 14.3\%$, and peak muscle activation by $28.2\%\sim54.0\%$.
This work provides new insights into adaptive exoskeleton assistance. However, because locomotion varies across individuals, the estimated torque cannot be fully adapted to different users. Future work should therefore explore personalized methods for estimating assistance torque.

    \section*{ACKNOWLEDGMENT}
    This work is funded by the National Natural Science Foundation of China (Grant 62622321, Grant 62473365, Grant U22A2056, and Grant 62373013) and the Beijing Natural Science Foundation (Grant L232021 and L242101).

    \bibliographystyle{plain}

\begin{thebibliography}{10}
\providecommand{\url}[1]{#1}
\csname url@samestyle\endcsname
\providecommand{\newblock}{\relax}
\providecommand{\bibinfo}[2]{#2}
\providecommand{\BIBentrySTDinterwordspacing}{\spaceskip=0pt\relax}
\providecommand{\BIBentryALTinterwordstretchfactor}{4}
\providecommand{\BIBentryALTinterwordspacing}{\spaceskip=\fontdimen2\font plus
\BIBentryALTinterwordstretchfactor\fontdimen3\font minus \fontdimen4\font\relax}
\providecommand{\BIBforeignlanguage}[2]{{%
\expandafter\ifx\csname l@#1\endcsname\relax
\typeout{** WARNING: IEEEtran.bst: No hyphenation pattern has been}%
\typeout{** loaded for the language `#1'. Using the pattern for}%
\typeout{** the default language instead.}%
\else
\language=\csname l@#1\endcsname
\fi
#2}}
\providecommand{\BIBdecl}{\relax}
\BIBdecl

\bibitem{slade2022personalizing}
P.~Slade, M.~J. Kochenderfer, S.~L. Delp, and S.~H. Collins, ``Personalizing exoskeleton assistance while walking in the real world,'' \emph{Nature}, vol. 610, no. 7931, pp. 277--282, 2022.

\bibitem{molinaro2024task}
D.~D. Molinaro, K.~L. Scherpereel, E.~B. Schonhaut, G.~Evangelopoulos, M.~K. Shepherd, and A.~J. Young, ``Task-agnostic exoskeleton control via biological joint moment estimation,'' \emph{Nature}, vol. 635, no. 8038, pp. 337--344, 2024.

\bibitem{2019emg}
K.~Gui, H.~Liu, and D.~Zhang, ``A practical and adaptive method to achieve emg-based torque estimation for a robotic exoskeleton,'' \emph{IEEE/ASME Transactions on Mechatronics}, vol.~24, no.~2, pp. 483--494, 2019.

\bibitem{quesada2026emg}
L.~Quesada, D.~Verdel, O.~Bruneau, B.~Berret, M.-A. Amorim, and N.~Vignais, ``Emg-to-torque models for exoskeleton assistance: a framework for the evaluation of in situ calibration,'' \emph{The International Journal of Robotics Research}, p. 02783649251414884, 2026.

\bibitem{emg2021}
C.~Caulcrick, W.~Huo, W.~Hoult, and R.~Vaidyanathan, ``Human joint torque modelling with mmg and emg during lower limb human-exoskeleton interaction,'' \emph{IEEE Robotics and Automation Letters}, vol.~6, no.~4, pp. 7185--7192, 2021.

\bibitem{2023id}
H.~Dinovitzer, M.~Shushtari, and A.~Arami, ``Accurate real-time joint torque estimation for dynamic prediction of human locomotion,'' \emph{IEEE Transactions on Biomedical Engineering}, vol.~70, no.~8, pp. 2289--2297, 2023.

\bibitem{2022id}
D.~D. Molinaro, I.~Kang, J.~Camargo, M.~C. Gombolay, and A.~J. Young, ``Subject-independent, biological hip moment estimation during multimodal overground ambulation using deep learning,'' \emph{IEEE Transactions on Medical Robotics and Bionics}, vol.~4, no.~1, pp. 219--229, 2022.

\bibitem{diraneyya2021inertial}
M.~M. Diraneyya, J.~Ryu, E.~Abdel-Rahman, and C.~T. Haas, ``Inertial motion capture-based whole-body inverse dynamics,'' \emph{Sensors}, vol.~21, no.~21, p. 7353, 2021.

\bibitem{liu2026exotraj}
X.-Y. Liu, G.~Li, L.~Sun, X.~Liang, and Z.-G. Hou, ``Exotraj: A general lower-limb exoskeleton assistance policy for complex environments,'' \emph{arXiv preprint arXiv:2606.16876}, 2026.

\bibitem{liu2025weight}
X.-Y. Liu, G.~Li, X.-H. Zhou, X.~Liang, and Z.-G. Hou, ``A weight-aware-based multisource unsupervised domain adaptation method for human motion intention recognition,'' \emph{IEEE Transactions on Cybernetics}, vol.~55, no.~7, pp. 3131--3143, 2025.

\bibitem{liu2026personalized}
X.-Y. Liu, G.~Li, W.~Wang, and Z.-G. Hou, ``Personalized lower-limb exoskeleton assistance via preference-based bayesian optimization,'' \emph{arXiv preprint arXiv:2608.09015}, 2026.

\bibitem{grillner2003motor}
S.~Grillner, ``The motor infrastructure: from ion channels to neuronal networks,'' \emph{Nature Reviews Neuroscience}, vol.~4, no.~7, pp. 573--586, 2003.

\bibitem{miall1993cerebellum}
R.~C. Miall, D.~J. Weir, D.~M. Wolpert, and J.~Stein, ``Is the cerebellum a smith predictor?'' \emph{Journal of Motor Behavior}, vol.~25, no.~3, pp. 203--216, 1993.

\bibitem{scott2004optimal}
S.~H. Scott, ``Optimal feedback control and the neural basis of volitional motor control,'' \emph{Nature Reviews Neuroscience}, vol.~5, no.~7, pp. 532--545, 2004.

\bibitem{wolpert2000computational}
D.~M. Wolpert and Z.~Ghahramani, ``Computational principles of movement neuroscience,'' \emph{Nature Neuroscience}, vol.~3, no.~11, pp. 1212--1217, 2000.

\bibitem{YAN2022439}
Y.~Yan, Z.~Chen, C.~Huang, L.~Chen, and Q.~Guo, ``Human-exoskeleton coupling dynamics in the swing of lower limb,'' \emph{Applied Mathematical Modelling}, vol. 104, pp. 439--454, 2022.

\bibitem{10716507}
L.~Chen, D.~Zhou, and Y.~Leng, ``A systematic review on rigid exoskeleton robot design for wearing comfort: Joint self-alignment, attachment interface, and structure customization,'' \emph{IEEE Transactions on Neural Systems and Rehabilitation Engineering}, vol.~32, pp. 3815--3827, 2024.

\bibitem{camargo2021comprehensive}
J.~Camargo, A.~Ramanathan, W.~Flanagan, and A.~Young, ``A comprehensive, open-source dataset of lower limb biomechanics in multiple conditions of stairs, ramps, and level-ground ambulation and transitions,'' \emph{Journal of Biomechanics}, vol. 119, p. 110320, 2021.

\bibitem{luo2024experiment}
S.~Luo, M.~Jiang, S.~Zhang, J.~Zhu, S.~Yu, I.~Dominguez~Silva, T.~Wang, E.~Rouse, B.~Zhou, H.~Yuk \emph{et~al.}, ``Experiment-free exoskeleton assistance via learning in simulation,'' \emph{Nature}, vol. 630, no. 8016, pp. 353--359, 2024.

\bibitem{molinaro2024estimating}
D.~D. Molinaro, I.~Kang, and A.~J. Young, ``Estimating human joint moments unifies exoskeleton control, reducing user effort,'' \emph{Science Robotics}, vol.~9, no.~88, p. eadi8852, 2024.

\bibitem{bird2026ultra}
C.~S. Bird, A.~P. Bo, W.~Lu, and T.~J. Dick, ``Ultra-wideband radar to measure in vivo muscle forces,'' \emph{Science Robotics}, vol.~11, no. 116, p. eaed5865, 2026.

\bibitem{9798696}
G.~Li, L.~Cheng, Z.~Gao, X.~Xia, and J.~Jiang, ``Development of an untethered adaptive thumb exoskeleton for delicate rehabilitation assistance,'' \emph{IEEE Transactions on Robotics}, vol.~38, no.~6, pp. 3514--3529, 2022.

\end{thebibliography}

    \newpage
\appendix

\subsection{System Description}

\begin{figure*}[b]
	\centering
	\includegraphics[width=0.96\textwidth]{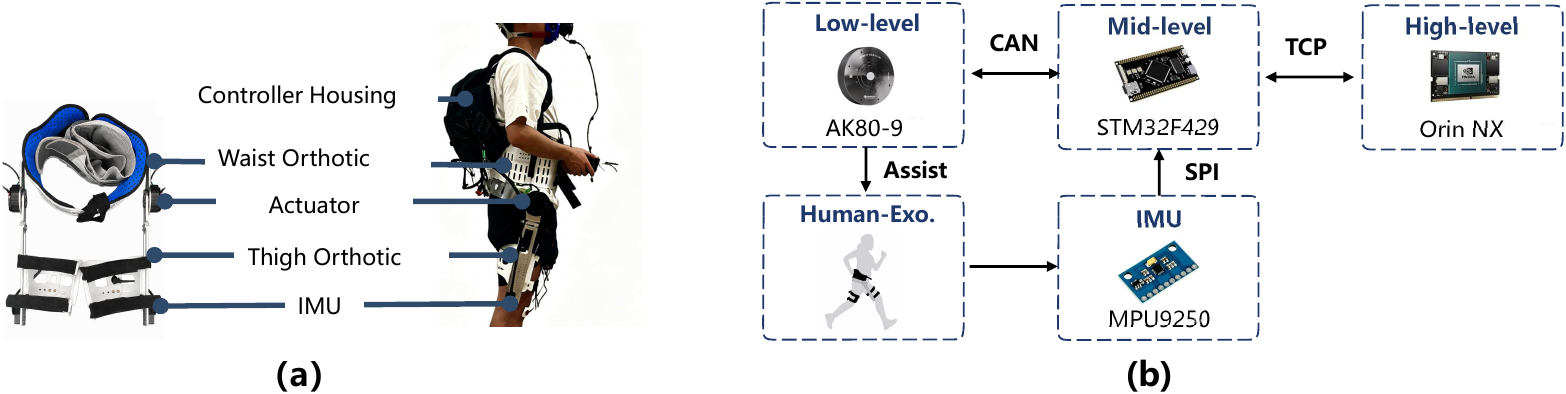}
	\caption{(a) shows the schematic diagram of a lower‑limb exoskeleton robot, and (b) gives the communication modes among various hardware devices. }. 
    \vspace{-0.4cm}
	\label{figa1}
\end{figure*}

Fig. \ref{figa1} shows the exoskeleton structure, hardware, and sensor communication architecture. As shown in Fig. \ref{figa1} (a), the hip exoskeleton weighs $6.8$ kg and comprises a waist harness, adjustable elastic belts, thigh support frames, a controller housing, and hip joint actuation modules, with customized orthoses securing it firmly to the user’s pelvis and thighs. A quasi-direct-drive actuator (AK80-9, T-Motor) provides torque assistance. Three IMUs (MPU9250) mounted on the bilateral thighs and lower torso capture locomotion kinematics. The NVIDIA Jetson Orin NX performs real-time neural-network torque inference, while the STM32F429 handles sensor sampling and motor drive control. Fig. \ref{figa1} (b) shows the communication architecture: TCP between the Jetson Orin NX and STM32, CAN for STM32–motor communication, and SPI for IMU data acquisition.

\begin{figure*}[b]
	\centering
	\includegraphics[width=0.98\textwidth]{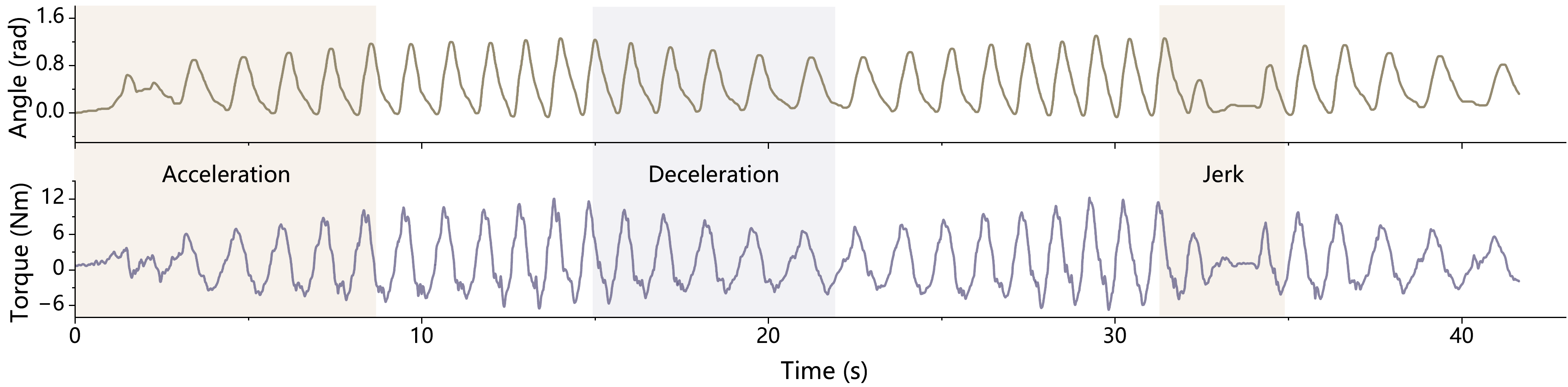}
	\caption{The results of the real-time torque prediction of the proposed method and related trajectory under rapid speed variations on the treadmill.}
    \vspace{-0.2cm}
	\label{figa2}
\end{figure*}

\subsection{Parameters Configuration}

This part presents detailed parameter settings for both the torque estimation model and the torque prediction network. For torque estimation based on the dynamic model, parameters of the dynamic model, \(\mathbf{M}\), \(\mathbf{C}\), \(\mathbf{G}\), error weight matrix \(\mathbf{Q}\), and torque weight matrix \(\mathbf{P}\) should be calibrated according to the specific exoskeleton prototype and control targets. The full configuration details are summarized in Table \ref{Configuration}.
For the torque prediction model, the network is based on the TCN model.
The input observation \(\mathbf{O}\) has dimensions \([250, 20]\) (\(T_o=250\)), and the output torque \(\mathbf{a}\) is a \([1,2]\) vector (\(T_p=1\)).
For training settings, the network is trained for $100$ epochs with a batch size of $512$ and an initial learning rate of \(1\times10^{-4}\). The entire network comprises approximately $1.5$ million parameters.
\begin{table}
	\normalsize
	\caption{Parameters Configuration of Torque Estimation and Prediction Model.}
	\label{Configuration}
	\centering
	\resizebox{8.8cm}{!}{
        \begin{tabular}{cc|cc}
\toprule[1pt] 
\multicolumn{2}{c|}{\textbf{Torque Estimation}}&\multicolumn{2}{c}{\textbf{Prediction Model }}\\
Parameter & Value & Parameter & Value \\
\midrule
Inertial mat. $\mathbf{M}$ & $0.0156\bb{I}$      
& Obs. horizon $T_o$ & $250$\\
Velocity mat. $\mathbf{C}$ & $\bb{0}$&
Future step $T_p$ & $1$\\
Gravity mat. $\mathbf{G}$ & $0.879 \sin (\bb{y})$&
Batch size & $512$\\
Coefficient $\mathbf{K}_1$ & $3 \bb{I}$&
Learning rate & $10^{-4}$\\
Error mat. $\mathbf{Q}$ & $\diag\{10,0.1\}\bb{I}$ &
Training epoch & $100$\\
Torque mat. $\mathbf{P}$ & $0.5\bb{I}$ &
Optimizer & Adam\\
\toprule[1pt] 
\end{tabular}
}
\end{table}

\subsection{Other Experimental Results}
Fig. \ref{figa2} illustrates the torque profiles generated by the proposed method during exoskeleton assistance under different treadmill speed conditions. As walking speed increases, the torque profiles exhibit progressively higher peak magnitudes and higher frequencies, indicating that the assistance is scaled according to gait cadence and intensity. Moreover, the system demonstrates rapid adaptability to velocity variations: when the user comes to a complete stop, the assistance torque decays to zero almost immediately with negligible latency, thereby avoiding unnecessary resistance or discomfort. These results confirm that the proposed method can effectively adapt to dynamically changing tasks involving rapid velocity variations, which is essential for practical exoskeleton applications in real-world locomotion.

\end{document}